\documentclass[journal]{IEEEtran}
\usepackage{etex}
\ifCLASSINFOpdf
\else
\fi
\usepackage{graphicx}
\usepackage{amsmath}
\usepackage{amsfonts}
\usepackage{amssymb}
\usepackage{url}
\usepackage{epsfig}
\usepackage{color}
\usepackage{epstopdf}
\usepackage{subcaption}

\usepackage{bm}
\usepackage[noend]{algpseudocode}
\usepackage[linesnumbered,ruled]{algorithm2e}

\usepackage{multicol}
\usepackage{longtable,ltxtable,booktabs}
\usepackage{lscape}
\usepackage{wrapfig}

\usepackage{pgfplots}
\DeclareMathOperator*{\argmin}{arg\,min}

\begin{document}
%
\title{Unsupervised Domain Adaptation: A Multi-task Learning-based Method}
%
%
%

\author{Jing~Zhang,~\IEEEmembership{Student Member,~IEEE,}
        Wanqing~Li,~\IEEEmembership{Senior~Member,~IEEE,}
        and~Philip~Ogunbona,~\IEEEmembership{Senior~Member,~IEEE}
\thanks{J. Zhang, W. Li, and P. Ogunbona are with the
Advanced Multimedia Research Lab, University of Wollongong, Wollongong, Australia. (e-mail: jz960@uowmail.edu.au; wanqing@uow.edu.au; philipo@uow.edu.au)}
\thanks{Manuscript received XXX; revised XXX.}}

%
%

\markboth{IEEE TRANSACTIONS ON CIRCUITS AND SYSTEMS FOR VIDEO TECHNOLOGY,~Vol.XXX, No.XXX, XXX}%
{Shell \MakeLowercase{\textit{et al.}}: Bare Demo of IEEEtran.cls for IEEE Journals}
%



\maketitle

\begin{abstract}
This paper presents a novel multi-task learning-based method for unsupervised domain adaptation. Specifically, the source and target domain classifiers are jointly learned by considering the geometry of target domain and the divergence between the source and target domains based on the concept of multi-task learning. Two novel algorithms are proposed upon the method using Regularized Least Squares and Support Vector Machines respectively. Experiments on both synthetic and real world cross domain recognition tasks have shown that the proposed methods outperform several state-of-the-art domain adaptation methods.
\end{abstract}


%
\IEEEpeerreviewmaketitle

\section{Introduction}
\IEEEPARstart{I}t is generally assumed that the training and test data are drawn from the same distribution in statistical learning theory. Unfortunately, this assumption does not hold in many applications. A well studied strategy to address this issue is domain adaptation~\cite{Ben-David2007,Pan2011,Long2014a} which employs previous labelled source domain data to boost the task in a new target domain with a few or even no labelled data.

In this paper, we focus on the problem of unsupervised domain adaptation in which source data are labelled, but available data in the target domain are unlabelled.
One approach to this problem is referred to as feature transformation-based domain adaptation~\cite{Pan2011,Long2013,Baktashmotlagh2013,Ghifary2016,Zhang2017}, which transforms the original feature into another space where the distributions of the two domains would be similar. A classifier can then be trained in the space for both source and target data. Another approach~\cite{Quanz2009,Long2014a}, referred to as classifier-based domain adaptation, aims at adapting a classifier directly by constraining the standard learning framework on the source labelled data to the unlabelled target data. However, there may not exist such a classifier that could perform well on both domains, especially when the domain shift is large. An alternative approach is to jointly learn two classifiers, one is optimized for the source domain and the other is optimized for the target domain. Yang et al.~\cite{Yang2007} proposed a method to learn two classifiers for the source and target domain respectively by assuming both the source and target data are labelled (i.e. supervised domain adaptation). Duan et al.~\cite{Duan2012a} propose a multi source domain adaptation method while the target domain labelled data are also assumed to be available. How to jointly learn source and target classifiers for unsupervised domain adaptation is a challenging problem despite the recent attempt in~\cite{Long2016} based on Residual Transfer Network (RTN).

This paper proposes a multi-task learning-based method for classifier-based domain adaptation. It aims at jointly learning the source and target classifiers without requiring labelled data in the target domain. Specifically, the target task is treated as an unsupervised clustering task by exploiting the intrinsic structure of unlabelled target data. In the meantime, the class information from the source domain is leveraged to assign the right class labels to the target clusters by taking the class distribution shift between the domains into consideration. The proposed method has been evaluated through comprehensive experiments on a synthetic dataset and real world cross domain visual recognition tasks. The experimental results demonstrate that the proposed method outperforms several state-of-the-art domain adaptation methods.

\section{Proposed Method}
This section presents the multi-task learning-based Unsupervised Domain Adaptation (mtUDA) method in detail. It begins with the definitions of terminologies. The source domain data denoted as $X_s\in{\mathbb{R}^{d\times n_s}}$ are draw from distribution $P_s(X_s)$ and the target domain data denoted as $X_t\in{\mathbb{R}^{d\times n_t}}$ are draw from distribution $P_t(X_t)$, where d is the dimension of the data instance, $n_s$ and $n_t$ are number of samples in source and target domain respectively. We focus on the unsupervised domain adaptation problem which assumes that there are sufficient labelled source domain data, $\mathcal{D}_s = \{(\mathbf{x}_i^s,y_i)\}_{i=1}^{n_s}$, $\mathbf{x}_i^s\in{\mathbb{R}^d}$, and unlabelled target domain data, $\mathcal{D}_t = \{(\mathbf{x}_j^t)\}_{j=1}^{n_t}$, $\mathbf{x}_j^t\in{\mathbb{R}^d}$, in the training stage. The feature spaces and label spaces between domains are assumed same: $\mathcal{X}_s = \mathcal{X}_t$ and $\mathcal{Y}_s = \mathcal{Y}_t$. However, due to the dataset shift, $P_s(X_s)\neq{P_t(X_t)}$ and $P_s(Y_s|X_s)\neq{P_t(Y_t|X_t)}$.

\subsection{Formulation}
\subsubsection{Regularized Risk Minimization}
Suppose there is no domain shift between source and target domains, the classifier learnt on the labelled source data can be applied on the target samples directly using the standard regularization based supervised learning algorithm in a Reproducing Kernel Hilbert Space (RKHS),
\begin{equation}
\min_{f\in{\mathcal{H}_K}} \sum_{i=1}^{n}\mathcal{L}(f(\mathbf{x}_i),y_i)+\gamma\|f\|_K^2
\end{equation} 
where $\mathcal{H}_K$ is an appropriately chosen RKHS, $K$ is a Mercer kernel, $\mathcal{L}$ denotes some loss functions, which can be squared loss $(y_i-f(\mathbf{x}_i))^2$ for RLS or hinge loss $max[0,1-y_if(\mathbf{x}_i)]$ for SVM, $\gamma$ is the shrinkage regularization parameter to regularize the complexity of learned model.

However, the source classifier is likely not optimal for the target domain due to the existence of domain shit. Hence, two different classifiers on source and target domain respectively are to learned with additional regularization on the two classifiers in a similar way to multi-task learning.

\subsubsection{Multi-task Learning}
The key idea of multi-task learning (MTL) is that the performances of the related tasks can be boosted by learning them jointly. In general, the MTL is based on the regularized risk minimization learning framework. In the context of domain adaptation, source and target tasks are two related tasks and the MTL formulation can be expressed as follows,
\begin{equation}
\begin{split}
\min_{f\in{\mathcal{H}_K}} &\frac{1}{n_s}\sum_{i=1}^{n_s}\mathcal{L}(f_s(\mathbf{x}_i^s),y_i)+\frac{1}{n_t}\sum_{j=1}^{n_t}\mathcal{L}(f_t(\mathbf{x}_j^t),y_j)\\
&+\gamma_A(\|f_s\|_K^2+\|f_t\|_K^2)+\gamma_M\Omega(f_s,f_t)
\end{split}\label{mtlf}
\end{equation}
where $\Omega$ is a regularization on the source and target classifiers, $\gamma_M$ is the classifier regularization parameter. In our problem, the source and target tasks are the same, suggesting that the label spaces between domains are identical. A simple regularization is $\Omega(f_s, f_t)=\|f_s-f_t\|_K^2$.

The MTL formulation in Eq.(\ref{mtlf}) requires labelled data in both domains. However, there are no labelled data in the target domain. Eq.(\ref{mtlf}) cannot be solved directly. In this paper, it is proposed to leverage the idea of manifold regularization~\cite{Belkin2006} to learn the intrinsic structure of the target domain to allow that the target task is treated as an unsupervised clustering task. 

\subsubsection{Multi-task Learning-based Unsupervised Domain Adaptation}
Since there are no labelled data in the target domain, the target risk minimization term in Eq.(\ref{mtlf}) is not computable. This paper proposes to replace the risk minimization term with an intrinsic regularization term to preserve the structure of the target data. Eq.(\ref{mtlf}) becomes
\begin{equation}
\begin{split}
\min_{f\in{\mathcal{H}_K}} &\frac{1}{n_s}\sum_{i=1}^{n_s}\mathcal{L}(f_s(\mathbf{x}_i^s),y_i)+\gamma_I\|f_t\|_I^2\\
&+\gamma_A(\|f_s\|_K^2+\|f_t\|_K^2)+\gamma_M\Omega(f_s,f_t)
\end{split}\label{rmtlf}
\end{equation}
where $\|f_t\|_I^2=\frac{1}{n_t^2}\sum_{i,j}(f_t(\mathbf{x}_i^t)-f_t(\mathbf{x}_j^t))^2W_{ij}=\frac{1}{n_t^2}tr(\mathbf{f}_t^TL\mathbf{f}_t)$ is the intrinsic manifold regularization for the target domain, $\mathbf{f}_t=[f_t(\mathbf{x}_1^t),...,f_t(\mathbf{x}_{n_t}^t)]^T$, $W_{ij}$ are edge weights in the adjacency graph (with p-nearest neighbours of each data point) of the target data, $L=D-W$ is the graph Laplacian, $D$ is a diagonal matrix given by $D_{ii}=\sum_{i,j=1}^{n_t}W_{ij}$, $\gamma_I$ is a manifold regularization parameter.

The use of multi-task regularization can remove additional conditions, such as orthogonal constraint, to avoid degenerate solutions as required in manifold regularization-based unsupervised learning~\cite{Belkin2006}. Note that the term $\|f_t\|_I^2$ is different from the manifold regularization term in~\cite{Long2014a}. It only regularizes the manifold structure in the target domain rather than the cross domain (all the data in source and target domains) as done in~\cite{Long2014a}, in order to reduce the distribution shift efficiently.

Eq.(\ref{rmtlf}) has considered the source risk minimization, task relatedness of the source and target domains, and target domain intrinsic structure in a multi-task learning framework. The target task here can be seen as an unsupervised clustering task. However, the ultimate goal is to assign class labels to target samples rather than just grouping them. Hence, the class information from the source domain needs to be leveraged. Since there is shift between the source and target domains and the manifold structure across the domains is not regularized, additional terms to reduce the class distribution shift are required to make sure that the target clusters are assigned with right class labels. In this paper, it is proposed to use Maximum Mean Discrepancy (MMD) criterion to regularize both the marginal and the conditional distribution shift~\cite{Long2013,Long2014a}.

The final objective function of the multi-task learning-based Unsupervised Domain Adaptation (mtUDA) method is
\begin{equation}
\begin{split}
\hspace{-1.5em}\min_{f_s,f_t\in{\mathcal{H}_K}} &\frac{1}{n_s}\sum_{i=1}^{n_s}\mathcal{L}(f_s(\mathbf{x}_i^s),y_i)+\gamma_I \|f_t\|_I^2+\gamma_A(\|f_s\|_K^2+\|f_t\|_K^2)\\
&+\gamma_M\Omega(f_s,f_t)+\gamma_D D(P_s,P_t)
\end{split}
\end{equation}
where $\gamma_D$ is the MMD regularization parameter,
\begin{equation}
\begin{split}
&D(P_s,P_t) = \|\frac{1}{n_s}\sum_{i=1}^{n_s}f_s(\mathbf{x}_i^s)-\frac{1}{n_t}\sum_{j=1}^{n_t}f_t(\mathbf{x}_j^t)\|_F^2\\
&+\sum_{c=1}^C\|\frac{1}{n^{(c)}_s}\sum_{\mathbf{x}_i^s\in{X_s^{(c)}}}f_s(\mathbf{x}_i^s)-\frac{1}{n^{(c)}_t}\sum_{\mathbf{x}_j^t\in{X_t^{(c)}}}f_t(\mathbf{x}_j^t)\|_F^2 \\
&= tr([\begin{matrix}
  \mathbf{f}_s^T & \mathbf{f}_t^T
 \end{matrix}]M\bigg[ \begin{matrix}
  \mathbf{f}_s\\
  \mathbf{f}_t
 \end{matrix} \bigg])
\end{split}
\end{equation}
is the MMD measure of joint distribution distance between source and target domains, $M = M_0+\sum_{c=1}^CM_c$, $(M_0)_{ij}=1/n_s^2$ if $\mathbf{x}_i,\mathbf{x}_j\in{\mathcal{D}_{s}}$, $(M_0)_{ij}=1/n_t^2$ if $\mathbf{x}_i,\mathbf{x}_j\in{\mathcal{D}_{t}}$, otherwise $(M_0)_{ij}=-1/n_sn_t$, 
\begin{equation*}
(M_c)_{ij} = 
  \begin{cases}
    \frac{1}{n_s^{(c)}n_s^{(c)}}      & \quad \text{if } \mathbf{x}_i,\mathbf{x}_j\in{\mathcal{D}_{s}^{(c)}} \\
    \frac{1}{n_t^{(c)}n_t^{(c)}}      & \quad \text{if } \mathbf{x}_i,\mathbf{x}_j\in{\mathcal{D}_{t}^{(c)}} \\
    \frac{-1}{n_s^{(c)}n_t^{(c)}}      & \quad \begin{cases} 
    \mathbf{x}_i\in{\mathcal{D}_{s}^{(c)}}, \mathbf{x}_j\in{\mathcal{D}_{t}^{(c)}} \\
    \mathbf{x}_j\in{\mathcal{D}_{s}^{(c)}}, \mathbf{x}_i\in{\mathcal{D}_{t}^{(c)}} \\
          \end{cases}\\ 0 &
\quad   \text{otherwise } \\
  \end{cases},
\label{eqt:M}
\end{equation*}
$\mathbf{f}_s=[f_s(\mathbf{x}_1),...,f_s(\mathbf{x}_{n_s})]^T$. Since there is no labelled data in the target domain, pseudo labels are obtained using some base classifiers (e.g. NN) trained on the source domain data in a similar way to~\cite{Long2014a}. Thus, the conditional distributions can be compared. The pseudo labels are iteratively updated after obtaining the adaptive classifier for the target data. It is worth emphasizing that the source and target classifiers are distinct, which is different from~\cite{Long2014a}.

In the following, two mtUDA algorithms with two different loss functions are presented, namely Regularized Least Squares and Support Vector Machines.

\subsection{Regularized Least Squares Algorithm} 
The Regularized Least Squares algorithm (denoted as mtUDA-RLS) can be expressed as,
\vspace{-0.5em}
\begin{equation}
\begin{split}
\min_{f_s,f_t\in{\mathcal{H}_K}} &\frac{1}{n_s}\sum_{i=1}^{n_s} (y_i-f_s(\mathbf{x}_i^s))^2+ \frac{\gamma_I}{n_t^2}tr(\mathbf{f}_t^TL\mathbf{f}_t)\\
&+\gamma_A(\|f_s\|_K^2+\|f_t\|_K^2)+\gamma_M \|f_s-f_t\|_K^2\\
&+\gamma_D tr([\begin{matrix}
  \mathbf{f}_s^T & \mathbf{f}_t^T
 \end{matrix}]M\bigg[ \begin{matrix}
  \mathbf{f}_s\\
  \mathbf{f}_t
 \end{matrix} \bigg])
\end{split}
\label{eqt:DARLS}
\end{equation}
Based on the Representer Theorem, the solution is an expansion of kernel functions over all the data:
\begin{equation}
\begin{split}
& f_s^{\star}(\mathbf{x}^s)=\sum_{i=1}^{n_s+n_t} \alpha_{s_i}^{\star}K(\mathbf{x}^s,\mathbf{x}_i),\\
& f_t^{\star}(\mathbf{x}^t)=\sum_{i=1}^{n_s+n_t} \alpha_{t_i}^{\star}K(\mathbf{x}^t,\mathbf{x}_i)
\label{eqt:fsft}
\end{split}
\end{equation}
Substituting them into Eq.(\ref{eqt:DARLS}), the following objective function can be obtained,
\begin{equation}
\begin{split}
(\boldsymbol{\alpha}_s^{\star},\boldsymbol{\alpha}_t^{\star})&=\hspace{-1.5em}\argmin_{\boldsymbol{\alpha}_s,\boldsymbol{\alpha}_t\in{\mathbb{R}^{(n_s+n_t) \times C}}}\frac{1}{n_s}tr((Y_s-\boldsymbol{\alpha}_s^TK_s^T)(Y_s-\boldsymbol{\alpha}_s^TK_s^T)^T)\\
&+ \frac{\gamma_I}{n_t^2}tr(\boldsymbol{\alpha}_t^TK_t^TLK_t\boldsymbol{\alpha}_t)+\gamma_A tr(\boldsymbol{\alpha}_s^TK\boldsymbol{\alpha}_s+\boldsymbol{\alpha}_t^TK\boldsymbol{\alpha}_t)\\
&+\gamma_M tr(\boldsymbol{\alpha}_s^TK\boldsymbol{\alpha}_s-\boldsymbol{\alpha}_s^TK\boldsymbol{\alpha}_t-\boldsymbol{\alpha}_t^TK\boldsymbol{\alpha}_s+\boldsymbol{\alpha}_t^TK\boldsymbol{\alpha}_t)\\
&+\gamma_D tr([\begin{matrix}
  \boldsymbol{\alpha}_s^TK_s^T & \boldsymbol{\alpha}_t^TK_t^T
 \end{matrix}]M\bigg[ \begin{matrix}
  K_s\boldsymbol{\alpha}_s\\
  K_t\boldsymbol{\alpha}_t
 \end{matrix} \bigg]) 
\end{split}
\label{eqt:DARLSalpha}
\end{equation}
where $Y_s\in \mathbb{R}^{C\times n_s}$ is a label matrix,  $y_s^c=1$ if $y_s(x)=c$, otherwise $y_s^c=0$ (mtUDA-RLS can be naturally applied to multi-class classification problem directly with this form of label matrix), $K = \Phi(X)^T\Phi(X)$, $K_s = \Phi(X_s)^T\Phi(X)$, and $K_t = \Phi(X_t)^T\Phi(X)$ are the kernel matrices, $X=[X_s, X_t]$ denotes all the source and target training samples, $\Phi(X)=[\phi(x_1),...,\phi(x_{n_s+n_t})]$ is the feature mappings to a space of a higher or even infinite dimension.

To simultaneously optimize $\boldsymbol{\alpha}_s$ and $\boldsymbol{\alpha}_t$, we write $[\begin{matrix}
  \boldsymbol{\alpha}_s^T & \boldsymbol{\alpha}_t^T
 \end{matrix}]$ as $\boldsymbol{\alpha}^T$. 
The objective function (Eq.(\ref{eqt:DARLSalpha})) can be solved efficiently in closed form.
Following~\cite{Belkin2006}, each trade-off coefficient is treated as a whole, e.g. $\hat{\gamma_I}=\frac{\gamma_In_s}{n_t^2}$, $\hat{\gamma_A}=\gamma_A n_s$, $\hat{\gamma_M}=\gamma_M n_s$, and $\hat{\gamma_D}=\gamma_D n_s$, when tuning the parameters.

\subsection{Support Vector Machines Algorithm}
The Support Vector Machines algorithm (denoted as mtUDA-SVM) can be formulated as,
\begin{equation}
\begin{split}
\min_{f_s,f_t\in{\mathcal{H}_K}} &\frac{1}{n_s}\sum_{i=1}^{n_s} \max(0,1-y_if_s(\mathbf{x}_i))_+\frac{\gamma_I}{n_t^2} tr(\mathbf{f}_t^TL\mathbf{f}_t)\\
&+\gamma_A(\|f_s\|_K^2+\|f_t\|_K^2)+\gamma_M \|f_s-f_t\|_K^2\\
&+\gamma_D tr([\begin{matrix}
  \mathbf{f}_s^T & \mathbf{f}_t^T
 \end{matrix}]M\bigg[ \begin{matrix}
  \mathbf{f}_s\\
  \mathbf{f}_t
 \end{matrix} \bigg])
\end{split}
\label{eqt:DASVM}
\end{equation} 
which can be solved using any SVM solver. Similar to mtUDA-RLS and~\cite{Belkin2006}, each trade-off coefficient is treated as a whole, e.g. $\hat{\gamma_I}=\frac{\gamma_I}{n_t^2}$, $\hat{\gamma_A}=\gamma_A$, $\hat{\gamma_M}=\gamma_M$, and $\hat{\gamma_D}=\gamma_D$.

For easily tuning the parameters, $M$ and $L$ matrices are normalized.

\section{Experiments}
\subsection{Set-ups}
\paragraph{Datasets}
The proposed mtUDA methods were evaluated on a synthetic dataset, real world digit recognition datasets and object recognition datasets. The synthetic data has two classes. The two classes of source data are generated by sampling from Gaussians centered at (0,2), and (2,0), while the two classes of target data are sampled from Gaussians centered at (-1,-1), and (2,0). 

For cross-domain hand-written digit recognition, MNIST~\cite{LeCun1998} and USPS~\cite{Hull1994} datasets were used. Data released by~\cite{Long2013,Long2014a} were used to construct a pair of cross-domain datasets USPS v.s. MNIST by randomly sampling 1,800 images in USPS and 2,000 images in MNIST. All images were uniformly rescaled to size 16$\times$16, and each image is represented by a feature vector encoding the gray-scale pixel values.

For object recognition, the methods were evaluated on two different datasets. One is Office-31 dataset studied by Saenko et al.~\cite{Saenko2010}, which contains three different domains: Amazon (images downloaded from online merchants), Webcam (low-resolution images by a web camera), DSLR (high-resolution images by a digital SLR camera). There are 31 classes of objects shared by the three domains, forming 6 pairs of cross domain tasks. $Decaf_7$ features~\cite{Donahue2014}~\footnote{\url{https://cs.stanford.edu/~jhoffman/domainadapt/}} were used. The other object dataset is the Office-Caltech-10 dataset released by Gong et al.~\cite{Gong2012}. This dataset is built upon the Office-31 dataset and contains images from four different domains: Amazon, Webcam, DSLR, and Caltech-256, where Caltech-256~\cite{Griffin2007} contains 256 object classes downloaded from Google images. Ten classes common to four datasets are selected to form 12 pairs of datasets. Two types of features are considered: $Decaf_7$ features~\cite{Donahue2014} and SURF descriptors~\footnote{\url{http://www-scf.usc.edu/~boqinggo/domainadaptation.html}}. 

\paragraph{Baselines and Settings}
The proposed methods are compared with the state-of-the-art unsupervised domain adaptation method, namely SA~\cite{Fernando2013}, JDA~\cite{Long2013}, JGSA~\cite{Zhang2017}, ARTL~\cite{Long2014a} (includes ARRLS and ARSVM), and RTN~\cite{Long2016}. SA, JDA, and JGSA are feature transformation-based methods, ARTL is a classifier-based method, and RTN is a joint feature adaptation and classifier adaptation method. The proposed methods are also compared with the No Adaptation baseline, which is the results obtained by the Nearest Neighbour classifier on the source domain data without adaptation. For all the compared methods, the parameters recommended by the original papers were used. For the proposed methods, $\hat{\gamma_M}=1$, $\hat{\gamma_A}=0.1$, $\hat{\gamma_I}=1$, $p=5$, and 1) $\hat{\gamma_D}=10$ for Office-31; 2) $\hat{\gamma_D}=1$ for other datasets. In fact, the proposed methods perform well on a wide range of parameter values based on the empirical results. The number of iterations is fixed to 10 since the algorithms generally converge within 10 iterations. All algorithms were evaluated in a fully transductive setup~\cite{Gong2013}.

\subsection{Results}
\begin{figure}
    \centering
    \vspace{-1em}
    \begin{subfigure}[t]{0.24\textwidth}
        \centering
        \includegraphics[scale=0.19]{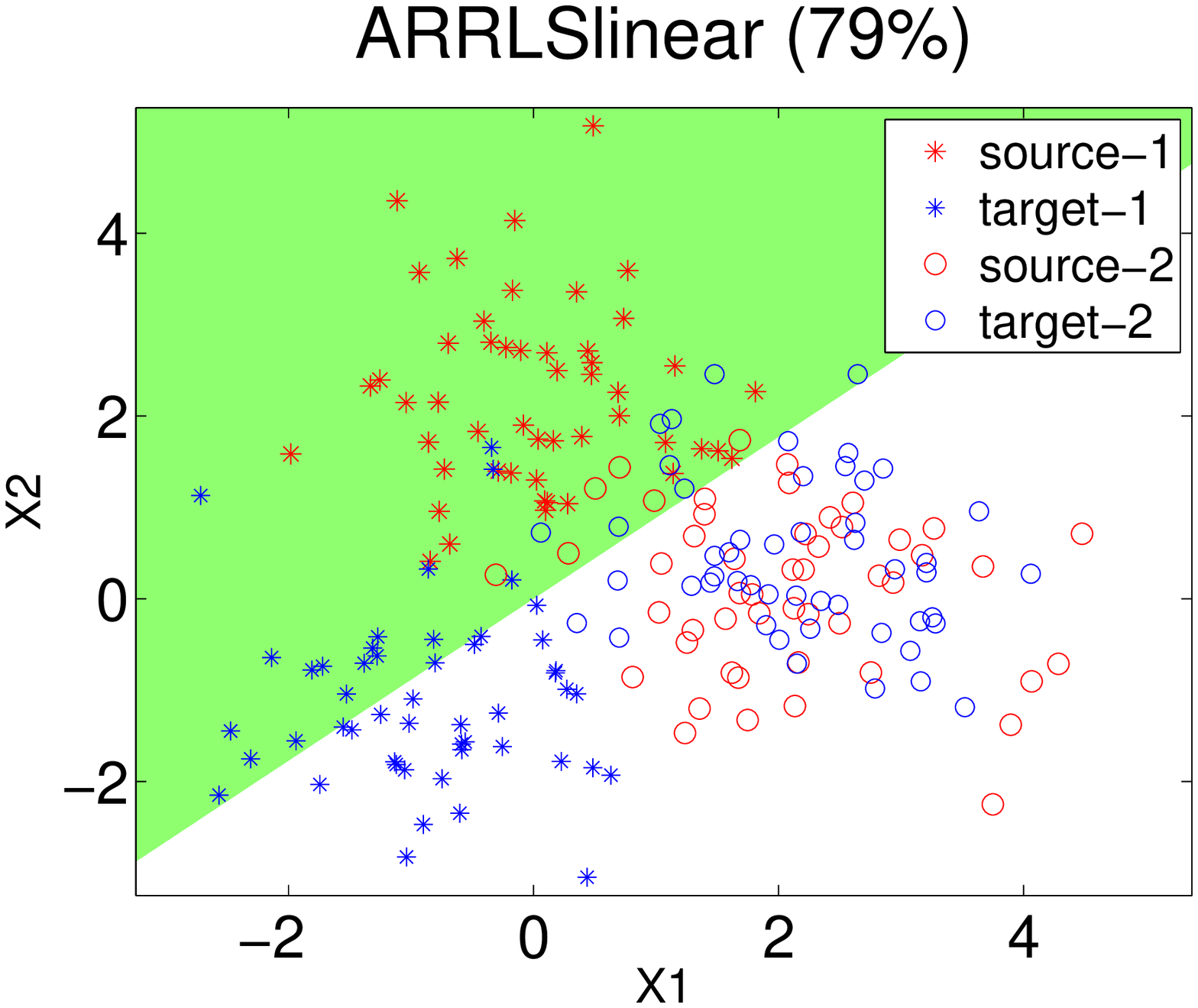}
\label{fig:DA2}
    \end{subfigure}%
~    
    \begin{subfigure}[t]{0.24\textwidth}
        \centering
        \includegraphics[scale=0.19]{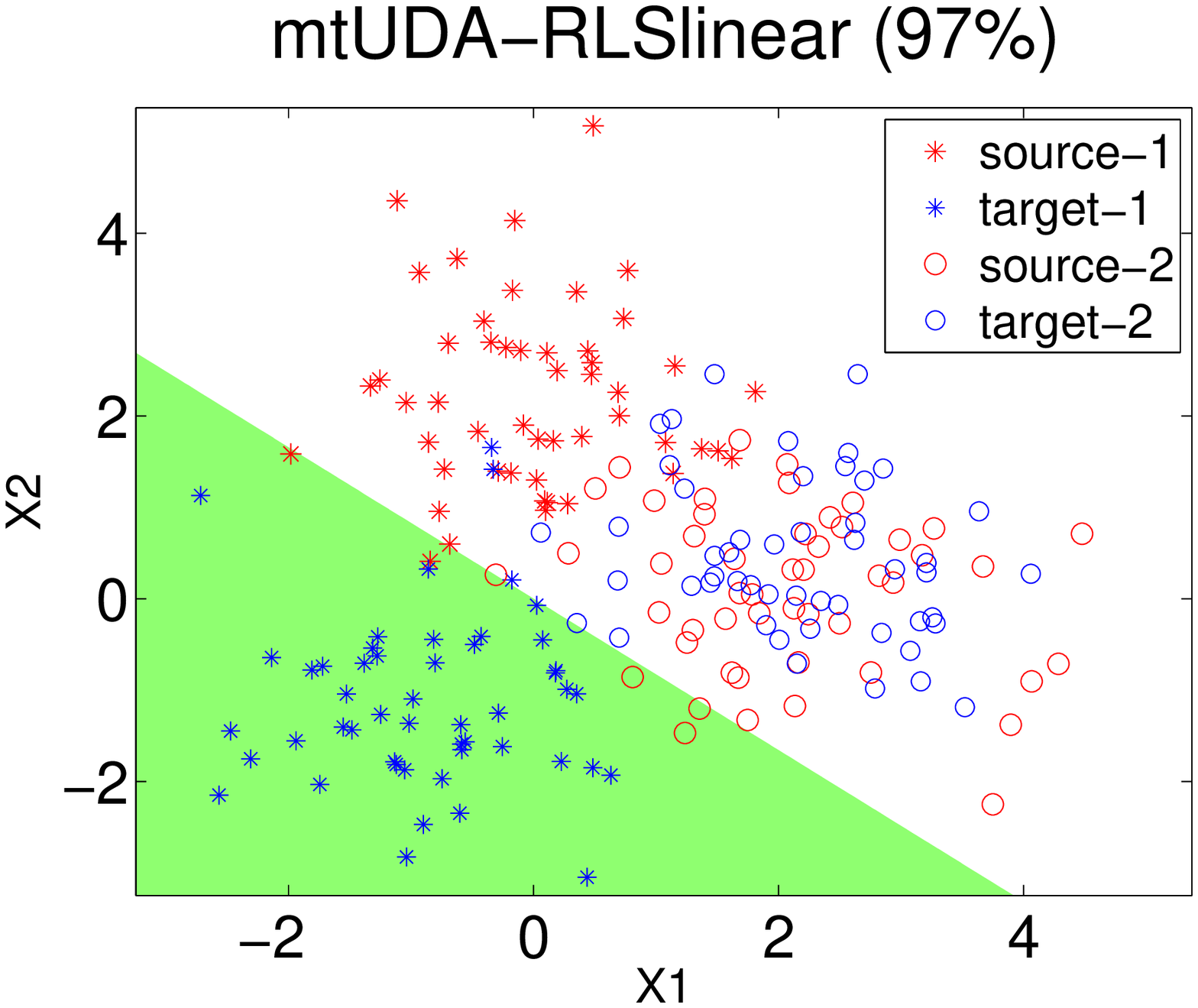}
\label{fig:DA4}
    \end{subfigure}
~
    \begin{subfigure}[t]{0.24\textwidth}
        \centering
        \includegraphics[scale=0.19]{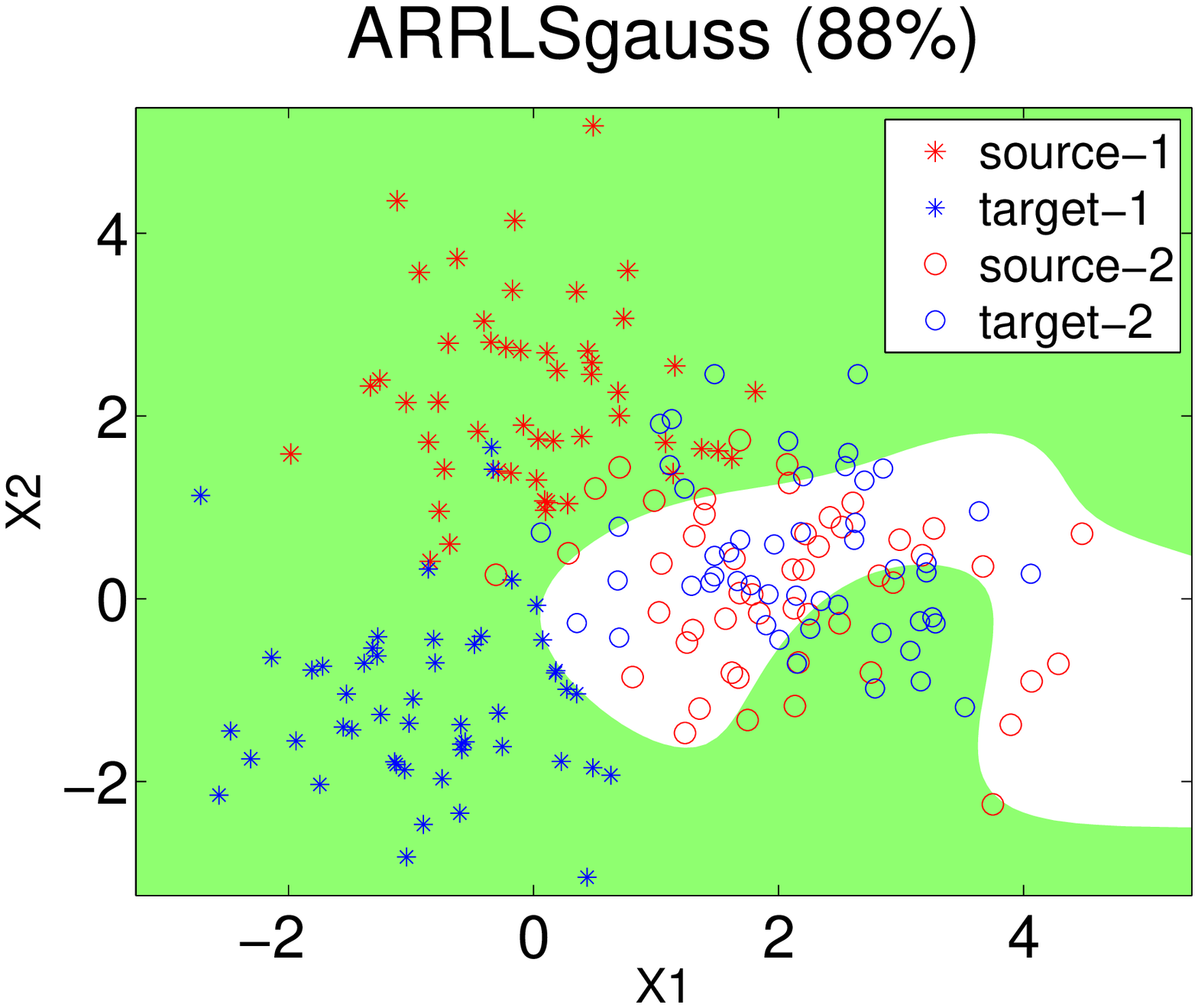}
\label{fig:DA6}
    \end{subfigure}%
~
    \begin{subfigure}[t]{0.24\textwidth}
        \centering
        \includegraphics[scale=0.19]{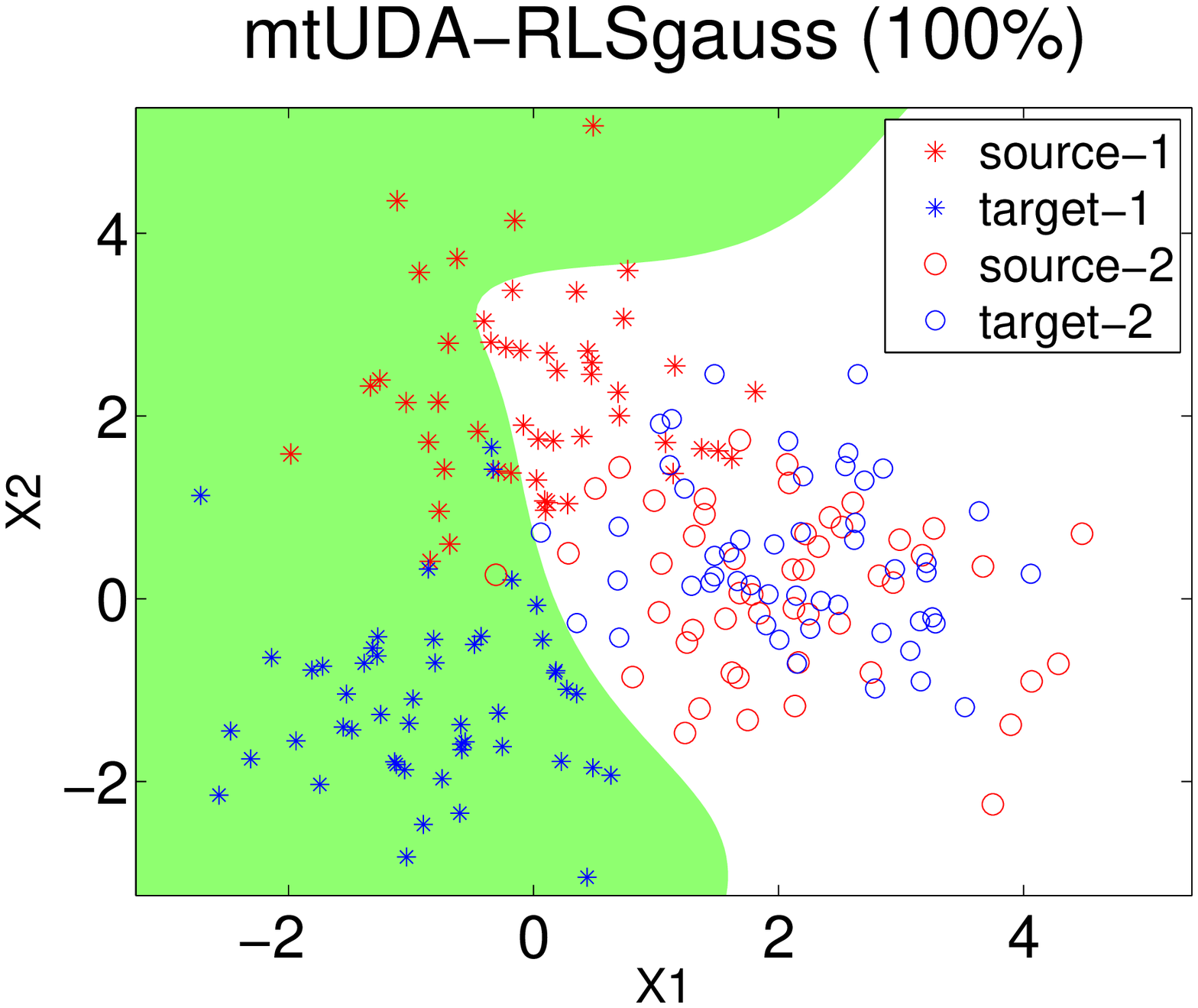}
\label{fig:DA8}
    \end{subfigure}
\vspace{-1.5em}
    \caption{Comparisons of ARRLS and the proposed mtUDA-RLS on the synthetic data.}
    \label{fig:toy}
\end{figure}

Figure~\ref{fig:toy} shows the comparison between  ARRLS~\cite{Long2014a} and the proposed mtUDA-RLS methods with linear and Gaussian kernels. The green shade represents the adaptive classifier. The ARRLS method tries to learn a unified classifier that can perform well in both domains, which does not exist or is hard to find when the domain shift is large. Since the goal of the adaptive classifier is to classify the target samples, it is not necessary to perform well in both domains. By contrast, our mtUDA-RLS method jointly learns two different but related classifiers for source and target domain respectively. Without the unified classifier constraint, the target domain adaptive classifier obtained by our mtUDA-RLS method performs much better than ARRLS on the target task.

\begin{table}[h]
\begin{small}
\begin{minipage}{0.47\textwidth}  
\caption{Accuracy on the digit dataset.}
  \label{tab:digit}
  \centering
  \begin{tabular}{m{3cm}m{1.8cm}m{1.8cm}m{0.45cm}}
  \toprule
  Datasets & MNIST$\rightarrow$USPS & USPS$\rightarrow$MNIST & Avg.\\
  \midrule
   No Adaptation & 65.9 & 44.7 & 55.3\\
   SA~\cite{Fernando2013} & 67.8 & 48.8 & 58.3\\
   JDA~\cite{Long2013} & 67.3 & 59.7 & 63.5\\
   JGSA~\cite{Zhang2017} & 80.4 & 68.2 & 74.3\\
   ARSVMlinear~\cite{Long2014a} & 67.8 & 57.2 & 62.5\\
   ARSVMGaussian~\cite{Long2014a} & 88.2 & 64.4 & 76.3\\
   ARRLSGaussian~\cite{Long2014a} & 88.8 & 67.7 & 78.2\\
  \midrule
   mtUDA-SVMlinear & 75.7 & 62.6 & 69.2\\
   mtUDA-SVMGaussian & 88.3 & 68.0 & 78.1\\
   mtUDA-RLSGaussian & \textbf{89.2} & \textbf{71.9} & \textbf{80.5}\\
  \bottomrule
  \end{tabular}
  \end{minipage}
\end{small}
\end{table}

\begin{table}[ht!]
\begin{center}
\begin{small}
  \caption{Accuracy on the Office-31 dataset with $decaf_7$ features.}
  \label{tab:office31}
  \centering
  \begin{tabular}{lm{0.4cm}m{0.4cm}m{0.4cm}m{0.4cm}m{0.4cm}m{0.4cm}m{0.4cm}}
    \toprule
    Datasets  & A$\rightarrow$W & A$\rightarrow$D & W$\rightarrow$A & W$\rightarrow$D & D$\rightarrow$A & D$\rightarrow$W & Avg. \\
    \midrule
    No Adaptation & 55.6 & 59.2 & 41.8 & 98.2 &  44.9 & 93.1 & 65.5 \\
    SA~\cite{Fernando2013} & 55.4 & 58.4 & 44.0 & 98.6 &  46.5 & 92.7 & 65.9 \\
    JDA~\cite{Long2013} & 56.9 & 56.6 & 45.0 & 98.0 &  47.0 & 94.1 & 66.3 \\
    JGSA~\cite{Zhang2017} & 60.4 & 65.3 & 54.0 & 97.2 &  51.5 & 96.4 & 70.8 \\
    ARRLSlinear~\cite{Long2014a} & 66.0 & 69.7 & 57.4 &  97.8 & 58.5 & 94.5 & 74.0\\
    ARRLSGaussian~\cite{Long2014a} & 68.3 & 70.3 & \textbf{60.2} &  98.4 & 61.5 & 95.2 & 75.6 \\
    RTN~\cite{Long2016} & 70.2 & 69.3 & 51.0 & \textbf{99.3} & 50.5 & \textbf{96.8} & 72.9 \\
    \midrule
    mtUDA-RLSlinear & 69.2 & 72.9 & 58.0 & 98.4 & 61.4 & 95.4 & 75.9 \\
    mtUDA-RLSGaussian & \textbf{70.4} & \textbf{73.1} & 58.8 & 98.4 & \textbf{63.1} & 96.0 & \textbf{76.6} \\
    \bottomrule
  \end{tabular}
\end{small}
\end{center}
\end{table}

\begin{table*}[ht!]
  \caption{Accuracy on the Office-Caltech-10 dataset with $decaf_7$ features.}
  \label{tab:OfficeCaldecaf}
  \centering
  \begin{small}
  \begin{tabular}{m{3cm}m{0.5cm}m{0.5cm}m{0.5cm}m{0.5cm}m{0.5cm}m{0.5cm}m{0.5cm}m{0.5cm}m{0.5cm}m{0.5cm}m{0.5cm}m{0.5cm}m{0.5cm}}
    \toprule
    Datasets     & C$\rightarrow$A & C$\rightarrow$W & C$\rightarrow$D & A$\rightarrow$C & A$\rightarrow$W & A$\rightarrow$D & W$\rightarrow$C & W$\rightarrow$A & W$\rightarrow$D & D$\rightarrow$C & D$\rightarrow$A & D$\rightarrow$W & Avg. \\
    \midrule
    No Adaptation & 90.5 & 77.6 & 82.2 & 83.6 & 74.2 & 86.0 & 77.4 & 80.2 & \textbf{100.0} & 79.5 & 86.9 & \textbf{99.7} & 84.8 \\
    SA~\cite{Fernando2013}  & 90.0 & 85.8 & 86.0 & 80.0 & 86.4 & 88.5 & 79.6 & 85.2 & 99.4 & 77.9 & 87.5 & 95.3 & 86.8 \\
    JDA~\cite{Long2013}  & 89.6 & 78.6 & 87.3 & 81.8 & 77.3 & 83.4 & 83.1 & 90.4 & \textbf{100.0} & 82.6 & 91.7 & 98.3 & 87.0 \\
    JGSA~\cite{Zhang2017}  & 90.9 & 86.8 & 89.8 & 85.6 & 84.1 & 88.5 & 87.4 & 92.0 & \textbf{100.0} & 86.6 & 92.8 & 99.0 & 90.3 \\
    ARSVMlinear~\cite{Long2014a} & 92.7 & 94.9 & 93.6 & 88.4 & 91.2 & 89.2 & 87.2 & \textbf{92.8} & 99.4 & 86.9 & \textbf{93.7} & 96.3 & 92.2 \\
    ARSVMGaussian~\cite{Long2014a} & 92.6 & \textbf{95.6} & 95.6 & 88.3 & 90.9 & 89.2 & 87.2 & \textbf{92.8} & 99.4 & 87.0 & 93.2 & 96.3 & 92.3 \\
    \midrule
    mtUDA-SVMlinear & \textbf{93.1} & 92.2 & \textbf{98.1} & \textbf{89.1} & 91.2 & \textbf{93.6} & \textbf{88.1} & 92.6 & 98.1 & \textbf{89.1} & \textbf{93.7} & \textbf{99.7} & \textbf{93.2} \\
    mtUDA-SVMGaussian & 93.0 & 91.9 & 97.5 & 88.3 & \textbf{91.5} & 93.0 & \textbf{88.1} & 92.5 & 97.5 & 88.8 & \textbf{93.7} & \textbf{99.7} & 92.9 \\
    \bottomrule
  \end{tabular}
  \end{small}
\end{table*}

\begin{table*}[ht!]
  \caption{Accuracy on the Office-Caltech-10 dataset with SURF features.}
  \label{tab:OfficeCalsurf}
  \centering
  \begin{small}
  \begin{tabular}{m{3cm}m{0.5cm}m{0.5cm}m{0.5cm}m{0.5cm}m{0.5cm}m{0.5cm}m{0.5cm}m{0.5cm}m{0.5cm}m{0.5cm}m{0.5cm}m{0.5cm}m{0.5cm}}
    \toprule
    Datasets    & C$\rightarrow$A & C$\rightarrow$W & C$\rightarrow$D & A$\rightarrow$C & A$\rightarrow$W & A$\rightarrow$D & W$\rightarrow$C & W$\rightarrow$A & W$\rightarrow$D & D$\rightarrow$C & D$\rightarrow$A & D$\rightarrow$W & Avg. \\
    \midrule
    No Adaptation & 36.0 & 29.2 & 38.2 & 34.2 & 31.2 & 35.7 & 28.8 & 31.6 & 84.7 & 29.6 & 28.3 & 83.7 & 40.9 \\
    SA~\cite{Fernando2013}  & 49.3 & 40.0 & 39.5 & 40.0 & 33.2 & 33.8 & \textbf{35.2} & 39.3 & 75.2 & \textbf{34.6} & \textbf{39.9} & 77.0 & 44.7 \\
    JDA~\cite{Long2013}  & 44.8 & 41.7 & 45.2 & 39.4 & 38.0 & 39.5 & 31.2 & 32.8 & 89.2 & 31.5 & 33.1 & 89.5 & 46.3 \\
    JGSA~\cite{Zhang2017}  & 51.5 & 45.4 & 45.9 & \textbf{41.5} & 45.8 & 47.1 & 33.2 & 39.9 & 90.5 & 29.9 & 38.0 & \textbf{91.9} & 50.0 \\
    ARRLSGaussian~\cite{Long2014a} & 50.7 & 43.1 & 46.5 & 40.9 & 40.7 & 42.7 & 31.5 & 38.7 & \textbf{91.1} & 30.5 & 32.6 & \textbf{91.9} & 48.4 \\
    \midrule
    mtUDA-RLSGaussian & \textbf{55.1} & \textbf{56.3} & \textbf{50.3} & 40.0 & \textbf{50.2} & \textbf{49.0} & 32.8 & \textbf{40.6} & 86.0 & 30.5 & 35.5 & 90.9 & \textbf{51.4} \\
    \bottomrule
  \end{tabular}
  \end{small}
\end{table*}

For the real world datasets, the comparison results in Tables~\ref{tab:digit}, \ref{tab:office31}, \ref{tab:OfficeCaldecaf}, \ref{tab:OfficeCalsurf} show that the proposed method outperforms the state-of-the-art domain adaptation methods on most of the datasets.
Based on the results, It is observed that all the domain adaptation methods outperform the No Adaptation results, which means that the domain shift indeed exists on these datasets and both feature transformation-based methods and classifier-based methods can reduce the shift to different degrees. Secondly, the classifier-based methods perform better than feature transformation-based methods in general on the evaluated datasets, which verifies that the two-step solution in the feature transformation-based methods may not be optimal. Thirdly, the proposed mtUDA methods outperforms both ARTL and RTN methods. As analysed before, ARTL assumes shared classifier between domains, which may not exist. RTN is deep learning based method which rely on data augmentation and carefully tuned parameters, and is prone to local minima.
Lastly, the comparison between the linear and Gaussian kernel versions of the proposed algorithms have shown that, for the digit recognition tasks, the Gaussian kernel outperforms linear kernel to a large degree, but the differences are not obvious on the object recognition tasks.
 
\subsection{Parameter Sensitivity}
\begin{figure}[ht!]
    \begin{subfigure}[t]{0.26\textwidth}
\pgfplotsset{grid style={dotted,very thin,lightgray},compat=1.5.1}
\resizebox{120pt}{90pt}{%
\begin{tikzpicture}
\begin{axis}[
xtick={1,2,3,4,5,6,7,8,9,10,11},
ytick={10,20,30,40,50,60,70,80,90,100},
ymin=20,ymax=81,
yticklabel style={/pgf/number format/precision=0},
xticklabels={0.01, 0.02, 0.05, 0.1, 0.2, 0.5, 1, 2, 5, 10, 100},
  xlabel=(a) $\hat{\gamma_M}$ value (Classifier Regularization),
  ylabel=Accuracy(\%),
  label style={font=\large},
  grid=both,
  legend pos= south west,
  legend columns=2,
  legend style={font=\fontsize{8}{8}\selectfont,/tikz/column 2/.style={
                column sep=5pt,
            },}]

                        
\addplot+[line width=0.25mm,solid,color=black,mark=square*,mark options={color=black}] table [x=gammaM, y=C->Wsurf10]{datasub.dat};
\addlegendentry{C$\rightarrow$Wsurf10}

\addplot+[line width=0.25mm,dashed,color =black,mark=square*,mark options={color=black}] table [x=gammaM, y=C->Wsurf10baseline]{datasub.dat};
\addlegendentry{C$\rightarrow$Wsurf10baseline}

\addplot+[line width=0.25mm,color =blue,mark=diamond*,mark options={color=blue}] table [x=gammaM, y=USPS->MNIST]{datasub.dat};
\addlegendentry{USPS$\rightarrow$MNIST}

\addplot+[line width=0.25mm,dashed,color =blue,mark=diamond*,mark options={color=blue}] table [x=gammaM, y=USPS->MNISTbaseline]{datasub.dat};
\addlegendentry{USPS$\rightarrow$MNISTbaseline}

\addplot+[line width=0.25mm,color=red,mark=*,mark options={color=red}] table [x=gammaM, y=A->Ddecaf31]{datasub.dat};
\addlegendentry{A$\rightarrow$Ddecaf31}

\addplot+[line width=0.25mm,dashed,color =red,mark=*,mark options={color=red}] table [x=gammaM, y=A->Ddecaf31baseline]{datasub.dat};
\addlegendentry{A$\rightarrow$Ddecaf31baseline}

\end{axis}
\end{tikzpicture}
}
\vspace{-1em}
    \end{subfigure}%
    \begin{subfigure}[t]{0.26\textwidth}

\pgfplotsset{grid style={dotted,very thin,lightgray},compat=1.5.1}
\resizebox{120pt}{90pt}{%
\begin{tikzpicture}
\begin{axis}[
xtick={1,2,3,4,5,6,7},
ytick={10,20,30,40,50,60,70,80,90,100},
ymin=20,ymax=81,
yticklabel style={/pgf/number format/precision=0},
xticklabels={0.001, 0.005, 0.01, 0.05, 0.1, 0.5, 1},
  xlabel=(b) $\hat{\gamma_A}$ value (Shrinkage Regularization),
  ylabel=Accuracy(\%),
  label style={font=\large},
  grid=both,  legend pos= south west,
  legend columns=2,
  legend style={font=\fontsize{8}{8}\selectfont,/tikz/column 2/.style={
                column sep=5pt,
            },}]

\addplot+[line width=0.25mm,solid,color=black,mark=square*,mark options={color=black}] table [x=gammaA, y=C->Wsurf10]{dataregu.dat};
\addlegendentry{C$\rightarrow$Wsurf10}

\addplot+[line width=0.25mm,dashed,color =black,mark=square*,mark options={color=black}] table [x=gammaA, y=C->Wsurf10baseline]{dataregu.dat};
\addlegendentry{C$\rightarrow$Wsurf10baseline}

\addplot+[line width=0.25mm,color =blue,mark=diamond*,mark options={color=blue}] table [x=gammaA, y=USPS->MNIST]{dataregu.dat};
\addlegendentry{USPS$\rightarrow$MNIST}

\addplot+[line width=0.25mm,dashed,color =blue,mark=diamond*,mark options={color=blue}] table [x=gammaA, y=USPS->MNISTbaseline]{dataregu.dat};
\addlegendentry{USPS$\rightarrow$MNISTbaseline}

\addplot+[line width=0.25mm,color=red,mark=*,mark options={color=red}] table [x=gammaA, y=A->Ddecaf31]{dataregu.dat};
\addlegendentry{A$\rightarrow$Ddecaf31}

\addplot+[line width=0.25mm,dashed,color =red,mark=*,mark options={color=red}] table [x=gammaA, y=A->Ddecaf31baseline]{dataregu.dat};
\addlegendentry{A$\rightarrow$Ddecaf31baseline}

\end{axis}
\end{tikzpicture}
}
\vspace{-1em}
    \end{subfigure}\\
    \begin{subfigure}[t]{0.26\textwidth}

\pgfplotsset{grid style={dotted,very thin,lightgray},compat=1.5.1}
\resizebox{120pt}{90pt}{%
\begin{tikzpicture}
\begin{axis}[
xtick={1,2,3,4,5,6,7,8,9,10,11,12},
ytick={10,20,30,40,50,60,70,80,90,100},
ymin=20,ymax=81,
yticklabel style={/pgf/number format/precision=0},
xticklabels={0.1,0.2,0.4,0.6,0.8,1,2,4,6,8,10},
  xlabel=(c) $\hat{\gamma_D}$ value (MMD Regularization),
  ylabel=Accuracy(\%),
  label style={font=\large},
  grid=both,  legend pos= south west,
  legend columns=2,
  legend style={font=\fontsize{8}{8}\selectfont,/tikz/column 2/.style={
                column sep=5pt,
            },}]


\addplot+[line width=0.25mm,solid,color=black,mark=square*,mark options={color=black}] table [x=gammaD, y=C->Wsurf10]{datammd.dat};
\addlegendentry{C$\rightarrow$Wsurf10}

\addplot+[line width=0.25mm,dashed,color =black,mark=square*,mark options={color=black}] table [x=gammaD, y=C->Wsurf10baseline]{datammd.dat};
\addlegendentry{C$\rightarrow$Wsurf10baseline}

\addplot+[line width=0.25mm,color =blue,mark=diamond*,mark options={color=blue}] table [x=gammaD, y=USPS->MNIST]{datammd.dat};
\addlegendentry{USPS$\rightarrow$MNIST}

\addplot+[line width=0.25mm,dashed,color =blue,mark=diamond*,mark options={color=blue}] table [x=gammaD, y=USPS->MNISTbaseline]{datammd.dat};
\addlegendentry{USPS$\rightarrow$MNISTbaseline}

\addplot+[line width=0.25mm,color=red,mark=*,mark options={color=red}] table [x=gammaD, y=A->Ddecaf31]{datammd.dat};
\addlegendentry{A$\rightarrow$Ddecaf31}

\addplot+[line width=0.25mm,dashed,color =red,mark=*,mark options={color=red}] table [x=gammaD, y=A->Ddecaf31baseline]{datammd.dat};
\addlegendentry{A$\rightarrow$Ddecaf31baseline}

\end{axis}
\end{tikzpicture}
}
\vspace{-1em}
    \end{subfigure}%
    \begin{subfigure}[t]{0.26\textwidth}
\pgfplotsset{grid style={dotted,very thin,lightgray},compat=1.5.1}
\resizebox{120pt}{90pt}{%
\begin{tikzpicture}
\begin{axis}[
xtick={1,2,3,4,5,6,7,8,9,10},
ytick={10,20,30,40,50,60,70,80,90,100},
ymin=20,ymax=81,
yticklabel style={/pgf/number format/precision=0},
xticklabels={0.01, 0.02, 0.05, 0.1, 0.2, 0.5, 1, 2, 5, 10},
  xlabel=(d) $\hat{\gamma_I}$ value (Manifold Regularization),
  ylabel=Accuracy(\%),
  label style={font=\large},
  grid=both,
  legend pos= south west,
  legend columns=2,
  legend style={font=\fontsize{8}{8}\selectfont,/tikz/column 2/.style={
                column sep=5pt,
            },}]

                        
\addplot+[line width=0.25mm,solid,color=black,mark=square*,mark options={color=black}] table [x=gammaI, y=C->Wsurf10]{datamani.dat};
\addlegendentry{C$\rightarrow$Wsurf10}

\addplot+[line width=0.25mm,dashed,color =black,mark=square*,mark options={color=black}] table [x=gammaI, y=C->Wsurf10baseline]{datamani.dat};
\addlegendentry{C$\rightarrow$Wsurf10baseline}

\addplot+[line width=0.25mm,color =blue,mark=diamond*,mark options={color=blue}] table [x=gammaI, y=USPS->MNIST]{datamani.dat};
\addlegendentry{USPS$\rightarrow$MNIST}

\addplot+[line width=0.25mm,dashed,color =blue,mark=diamond*,mark options={color=blue}] table [x=gammaI, y=USPS->MNISTbaseline]{datamani.dat};
\addlegendentry{USPS$\rightarrow$MNISTbaseline}

\addplot+[line width=0.25mm,color=red,mark=*,mark options={color=red}] table [x=gammaI, y=A->Ddecaf31]{datamani.dat};
\addlegendentry{A$\rightarrow$Ddecaf31}

\addplot+[line width=0.25mm,dashed,color =red,mark=*,mark options={color=red}] table [x=gammaI, y=A->Ddecaf31baseline]{datamani.dat};
\addlegendentry{A$\rightarrow$Ddecaf31baseline}

\end{axis}
\end{tikzpicture}
}
\vspace{-1.8em}
    \end{subfigure}%
\vspace{-0.5em}
\caption{Parameter sensitivity of the proposed methods on different datasets}
\vspace{-0.5em}
\label{fig:params}
\end{figure}
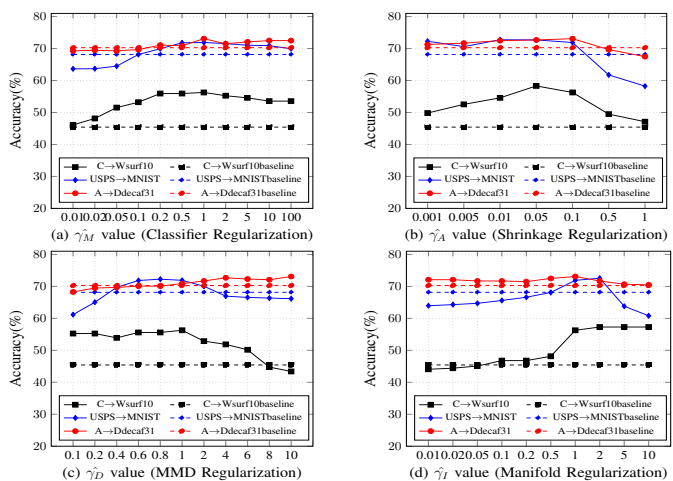

Experiments were conducted on the C$\rightarrow$W (Office-Caltech-10 dataset with SURF descriptor), USPS$\rightarrow$MNIST, and A$\rightarrow$D (Office-31 dataset with $decaf_7$ feature) datasets to study the sensitivity of the proposed algorithms to their parameters (Figure~\ref{fig:params}). The solid lines are the accuracies obtained by mtUDA-RLS with Gaussian kernel, and the dashed lines are the results obtained by the best baseline methods on each dataset. It can be seen that a wide range of values can be chosen to obtain satisfactory performances. $\hat{\gamma_M}$ regularizes the similarity between source and target classifiers. If $\hat{\gamma_M}$ is too small, the source class information cannot be transferred to the target domain, if $\hat{\gamma_M}\rightarrow \infty$ the source and target classifiers tend to be the same, which may not be desirable. $\hat{\gamma_A}$ controls the complexity of the classifiers. A small $\hat{\gamma_A}$ would lead to overfitting while a too large $\hat{\gamma_A}$ leads that the models cannot fit the data. $\hat{\gamma_D}$ controls the degree of distribution shift. Though the larger $\hat{\gamma_D}$ will lead to smaller distribution shift, a too large value will cancel out other regularizations. $\hat{\gamma_I}$ regularizes the geometry structure of target domain. If $\hat{\gamma_I}$ is too small, the target domain structure is not preserved, but if $\hat{\gamma_I}$ is too large the source class information is discarded.
\vspace{-0.5em}

\section{Conclusion}
This paper presents a novel multi-task learning-based unsupervised domain adaptation method. It relaxes the single classifier assumption in the conventional classifier-based unsupervised domain adaptation and proposes to jointly optimize source and target classifiers by considering the manifold structure of target domain and the distribution divergence between the domains. Experimental results on both synthetic and real world cross domain recognition datasets have shown the effectiveness of the proposed method.

\ifCLASSOPTIONcaptionsoff
  \newpage
\fi

\bibliographystyle{IEEEtran} 
\bibliography{CrossDataset}

\begin{thebibliography}{10}
\providecommand{\url}[1]{#1}
\csname url@samestyle\endcsname
\providecommand{\newblock}{\relax}
\providecommand{\bibinfo}[2]{#2}
\providecommand{\BIBentrySTDinterwordspacing}{\spaceskip=0pt\relax}
\providecommand{\BIBentryALTinterwordstretchfactor}{4}
\providecommand{\BIBentryALTinterwordspacing}{\spaceskip=\fontdimen2\font plus
\BIBentryALTinterwordstretchfactor\fontdimen3\font minus
  \fontdimen4\font\relax}
\providecommand{\BIBforeignlanguage}[2]{{%
\expandafter\ifx\csname l@#1\endcsname\relax
\typeout{** WARNING: IEEEtran.bst: No hyphenation pattern has been}%
\typeout{** loaded for the language `#1'. Using the pattern for}%
\typeout{** the default language instead.}%
\else
\language=\csname l@#1\endcsname
\fi
#2}}
\providecommand{\BIBdecl}{\relax}
\BIBdecl

\bibitem{Ben-David2007}
S.~Ben-David, J.~Blitzer, K.~Crammer, F.~Pereira \emph{et~al.}, ``Analysis of
  representations for domain adaptation,'' \emph{Advances in neural information
  processing systems}, vol.~19, p. 137, 2007.

\bibitem{Pan2011}
S.~J. Pan, I.~W. Tsang, J.~T. Kwok, and Q.~Yang, ``Domain adaptation via
  transfer component analysis,'' \emph{IEEE Transactions on Neural Networks},
  vol.~22, no.~2, pp. 199--210, 2011.

\bibitem{Long2014a}
M.~Long, J.~Wang, G.~Ding, S.~J. Pan, and S.~Y. Philip, ``Adaptation
  regularization: A general framework for transfer learning,'' \emph{IEEE
  Transactions on Knowledge and Data Engineering}, vol.~26, no.~5, pp.
  1076--1089, 2014.

\bibitem{Long2013}
M.~Long, J.~Wang, G.~Ding, J.~Sun, and P.~Yu, ``Transfer feature learning with
  joint distribution adaptation,'' in \emph{Proc. IEEE International Conference
  on Computer Vision}.\hskip 1em plus 0.5em minus 0.4em\relax IEEE, 2013, pp.
  2200--2207.

\bibitem{Baktashmotlagh2013}
M.~Baktashmotlagh, M.~T. Harandi, B.~C. Lovell, and M.~Salzmann, ``Unsupervised
  domain adaptation by domain invariant projection,'' in \emph{Proc. IEEE
  International Conference on Computer Vision}, 2013, pp. 769--776.

\bibitem{Ghifary2016}
M.~Ghifary, D.~Balduzzi, W.~B. Kleijn, and M.~Zhang, ``Scatter component
  analysis: A unified framework for domain adaptation and domain
  generalization,'' \emph{IEEE Transactions on Pattern Analysis and Machine
  Intelligence}, vol.~PP, no.~99, pp. 1--1, 2016.

\bibitem{Zhang2017}
J.~Zhang, W.~Li, and P.~Ogunbona, ``Joint geometrical and statistical alignment
  for visual domain adaptation,'' in \emph{Proc. IEEE Conference on Computer
  Vision and Pattern Recognition}, 2017.

\bibitem{Quanz2009}
B.~Quanz and J.~Huan, ``Large margin transductive transfer learning,'' in
  \emph{Proc. 18th ACM conference on Information and knowledge
  management}.\hskip 1em plus 0.5em minus 0.4em\relax ACM, 2009, pp.
  1327--1336.

\bibitem{Yang2007}
J.~Yang, R.~Yan, and A.~G. Hauptmann, ``Cross-domain video concept detection
  using adaptive svms,'' in \emph{Proc. 15th ACM international conference on
  Multimedia}.\hskip 1em plus 0.5em minus 0.4em\relax ACM, 2007, pp. 188--197.

\bibitem{Duan2012a}
L.~Duan, D.~Xu, and I.~W.-H. Tsang, ``Domain adaptation from multiple sources:
  A domain-dependent regularization approach,'' \emph{IEEE Transactions on
  Neural Networks and Learning Systems}, vol.~23, no.~3, pp. 504--518, 2012.

\bibitem{Long2016}
M.~Long, H.~Zhu, J.~Wang, and M.~I. Jordan, ``Unsupervised domain adaptation
  with residual transfer networks,'' in \emph{Advances in Neural Information
  Processing Systems}, 2016, pp. 136--144.

\bibitem{Belkin2006}
M.~Belkin, P.~Niyogi, and V.~Sindhwani, ``Manifold regularization: A geometric
  framework for learning from labeled and unlabeled examples,'' \emph{Journal
  of machine learning research}, vol.~7, no. Nov, pp. 2399--2434, 2006.

\bibitem{LeCun1998}
Y.~Lecun, L.~Bottou, Y.~Bengio, and P.~Haffner, ``Gradient-based learning
  applied to document recognition,'' \emph{Proceedings of the IEEE}, vol.~86,
  no.~11, pp. 2278--2324, Nov 1998.

\bibitem{Hull1994}
J.~J. Hull, ``A database for handwritten text recognition research,''
  \emph{IEEE Transactions on pattern analysis and machine intelligence},
  vol.~16, no.~5, pp. 550--554, 1994.

\bibitem{Saenko2010}
K.~Saenko, B.~Kulis, M.~Fritz, and T.~Darrell, ``Adapting visual category
  models to new domains,'' in \emph{Computer Vision--ECCV 2010}.\hskip 1em plus
  0.5em minus 0.4em\relax Springer, 2010, pp. 213--226.

\bibitem{Donahue2014}
J.~Donahue, Y.~Jia, O.~Vinyals, J.~Hoffman, N.~Zhang, E.~Tzeng, and T.~Darrell,
  ``Decaf: A deep convolutional activation feature for generic visual
  recognition,'' in \emph{Proc. 31st International Conference on Machine
  Learning}, 2014, pp. 647--655.

\bibitem{Gong2012}
B.~Gong, Y.~Shi, F.~Sha, and K.~Grauman, ``Geodesic flow kernel for
  unsupervised domain adaptation,'' in \emph{Proc. IEEE Conference on Computer
  Vision and Pattern Recognition}.\hskip 1em plus 0.5em minus 0.4em\relax IEEE,
  2012, pp. 2066--2073.

\bibitem{Griffin2007}
G.~Griffin, A.~Holub, and P.~Perona, ``Caltech-256 object category dataset,''
  Tech. Rep., 2007.

\bibitem{Fernando2013}
B.~Fernando, A.~Habrard, M.~Sebban, and T.~Tuytelaars, ``Unsupervised visual
  domain adaptation using subspace alignment,'' in \emph{Proc. IEEE
  International Conference on Computer Vision}.\hskip 1em plus 0.5em minus
  0.4em\relax IEEE, 2013, pp. 2960--2967.

\bibitem{Gong2013}
B.~Gong, K.~Grauman, and F.~Sha, ``Connecting the dots with landmarks:
  Discriminatively learning domain-invariant features for unsupervised domain
  adaptation,'' in \emph{Proc. 30th International Conference on Machine
  Learning}, 2013, pp. 222--230.

\end{thebibliography}

\end{document}